%% file: manuscript.tex
\newcounter{IncludeFigures}
\newcounter{IncludeTables}
\newcounter{IncludeSupplementaryFigures}
\documentclass[review,12pt,authoryear,compress]{elsarticle}

\usepackage{amsmath,amssymb} 
\usepackage{graphicx} 
\usepackage{booktabs} 
\usepackage{adjustbox}
\usepackage{rotating}
\usepackage{siunitx} 
\usepackage{chemformula} 
\usepackage{multirow} 

\usepackage{hyperref} 
\usepackage{cleveref}
\crefname{figure}{Figure}{Figures}       
\Crefname{figure}{Figure}{Figures}       
\usepackage{enumitem} 
\usepackage[utf8]{inputenc}
\usepackage{soul}      

\hypersetup{
  colorlinks=true, 
  linkcolor=blue, 
  citecolor=blue, 
  urlcolor=blue, 
  pdfborder={0 0 0}, 
  pdfauthor={}, 
  pdfsubject={}, 
  pdfkeywords={}, 
  pdfcreator={}, 
  pdfproducer={} 
}

\journal{}

\makeatletter
\def\ps@pprintTitle{%
  \let\@oddhead\@empty
  \let\@evenhead\@empty
  \let\@oddfoot\@empty
  \let\@evenfoot\@empty
  \def\footnoterule{}
}
\makeatother

\begin{document}

\begin{frontmatter}

\title{A neural recommender system leveraging transfer learning for property prediction of ionic liquids}

\author[a]{Sahil Sethi}
\author[a,b]{Kai Sundmacher}
\author[b]{Caroline Ganzer\corref{cor1}}

\affiliation[a]{organization={Otto von Guericke University Magdeburg, Chair of Process Systems Engineering}, 
               addressline={Universitätsplatz 2}, 
               city={Magdeburg}, 
               postcode={39106}, 
               country={Germany}}

\affiliation[b]{organization={Max Planck Institute for Dynamics of Complex Technical Systems, Department of Process Systems Engineering}, 
               addressline={Sandtorstr. 1}, 
               city={Magdeburg}, 
               postcode={39106}, 
               country={Germany}}

\cortext[cor1]{Corresponding author: cganzer@mpi-magdeburg.mpg.de}

\begin{abstract}
Ionic liquids (ILs) have emerged as versatile replacements for traditional solvents because their physicochemical properties can be precisely tailored to various applications. However, accurately predicting key thermophysical properties remains challenging due to the vast chemical design space and the limited availability of experimental data. In this study, we present a data-driven transfer learning framework combined with a neural recommender system (NRS) to enable reliable property prediction for ILs using sparse experimental datasets. The approach involves a two-stage process: first, pre-training NRS models on COSMO-RS-based simulated data at fixed temperature and pressure,  and second, fine-tuning simple feedforward neural networks with experimental data at varying temperatures and pressures. In this work, five essential IL properties are considered: density, viscosity, surface tension, heat capacity, and melting point. We find that the framework supports both within-property and cross-property knowledge transfer. Notably, pre-trained models for density, viscosity, and heat capacity are used to fine-tune models for all five target properties, achieving improved performance by a substantial margin for four of them. The model exhibits robust extrapolation to previously unseen ILs. Moreover, the final trained models enable property prediction for over 700,000 IL combinations, offering a scalable solution for IL screening in process design. This work highlights the effectiveness of combining simulated data and transfer learning to overcome sparsity in the experimental data.

\end{abstract}


\end{frontmatter}

\textbf{Highlights}
\begin{itemize}
\item We combine a neural recommender system with transfer learning to predict properties of ionic liquids from sparse experimental data.
\item Pre-training on simulation data from COSMO-RS boosts model performance.
\item Proposed strategy enables extrapolation with high accuracy for 700{,}000+ ionic liquids.
\end{itemize}

\section{Introduction}
Ionic liquids (ILs) are a class of salts that remain in the liquid state at or near room temperature and have gained attention as potential alternatives to conventional solvents \citep{Brennecke2001, Welton1999}. Their favorable properties, including low vapor pressure, high thermal stability, and a wide liquid temperature range, make them suitable for a variety of process engineering applications \citep{Patel2012}. In addition, ILs are structurally versatile and their physicochemical properties can be adjusted by selecting different combinations of cations and anions. This tunability offers the opportunity to design ILs tailored to specific process requirements. However, practical use of ILs requires knowledge of key properties such as density, viscosity, surface tension, heat capacity, and melting point. Experimentally determining these properties for the vast number of potential ILs is expensive and time-consuming. Therefore, predictive modeling has become an essential tool for screening ILs and guiding their selection in process design.

In the context of designing and selecting ILs for chemical processes, several fundamental physical properties play a critical role in determining their practical applicability and process performance. Density is essential for material and energy balances, equipment design, and phase behavior prediction \citep{Birdi2016}. Viscosity influences the flow behavior of ILs, affecting mass and heat transfer, mixing, and pumping requirements \citep{Shen2019}. Surface tension governs interfacial interactions and is especially relevant in applications involving liquid–liquid extraction, wetting, or emulsification \citep{Baskin2022}. Heat capacity reflects a fluid’s ability to store thermal energy and is essential for thermal management, energy integration, and accurate energy balance calculations in process modeling. Finally, melting point sets the lower operational limit for ILs and is used to exclude candidates that may solidify under working conditions. Reliable knowledge or prediction of these properties is vital for effective process modeling, simulation, optimization, and making informed process-related decisions. However, due to the enormous variety of possible IL structures, identifying suitable candidates with the right combination of properties remains a complex and computationally intensive task \citep{Wasserscheid2000, Stark2000}.

Traditionally, the identification of suitable ionic liquids for process applications has relied heavily on experimental screening, which is often slow, costly, and restricted to a narrow set of known IL structures. As a result, a large portion of the IL design space remains unexplored. To overcome these limitations, computational approaches have been increasingly used to estimate key thermophysical properties of ILs \citep{Sethi2025, Sengers2000}. Classical methods such as equations of state (EoS) \citep{Turner2003} and group contribution methods (GCMs) \citep{Gani1983} have been widely applied to predict properties like density, viscosity, and heat capacity. However, these approaches face important limitations. Equations of state often require parameter fitting and may not be generalizable to new IL structures, while GCMs depend on the availability and quality of group parameters, which are sometimes missing or inaccurate for novel ionic liquids. As a result, both techniques may lack the flexibility and precision needed to predict a broad set of IL properties across a wide structural range.

One of the widely adopted strategies for predicting the properties of ILs is the quantitative structure–property relationship (QSPR) approach. This method builds statistical models that associate target properties with structural features of ILs, typically encoded through molecular descriptors \citep{Sepehri2020, RybinskaFryca2020}. With the growth of publicly available IL property databases such as PubChem \citep{kim2025} and ILThermo \citep{dong2007}, a wide variety of descriptors ranging from simple topological indices to complex quantum chemical parameters have been used to represent ILs. In recent years, machine learning techniques have increasingly been employed within QSPR frameworks, enhancing their flexibility and predictive power for modeling diverse IL properties \citep{Cao2018, Chen2024, Sun2021, Ding2021, Wu2024, Baskin2022b, Mohan2024}. However, these descriptor-based models still depend heavily on the quality and completeness of the descriptors themselves, which are often designed manually and tailored to specific tasks \citep{Han2011, Rybinska2016, Xia2023, Lotfi2021}. This limits their ability to generalize across structurally diverse IL families. Graph neural networks (GNNs) have been explored for IL property prediction, leveraging molecular graph representations to capture structural relationships with no reliance on traditional molecular descriptors \citep{sanchez2026graph, rittig2023graph, qin2023capturing}. However, GNNs often require large datasets to effectively learn these representations, limiting their applicability when experimental data is scarce. As an alternative, matrix completion methods based on neural networks have emerged \citep{Chen2021, Zhang2022, Ramlatchan2018, Athey2021, Fan2018, Fan2017}. These methods treat the IL property dataset as a matrix with missing entries and aim to infer the unknown values by exploiting correlations between ILs and between properties. This approach eliminates the need for explicit descriptors and instead learns latent patterns directly from available data. However, such models require a sufficiently large amount of labeled data to train effectively and to predict the properties of ILs unseen in the training step. \citep{Chen2021, Zhang2022}. In cases of limited experimental data, they often struggle with overfitting or fail to capture meaningful structure-property relationships.

To address the challenge of data sparsity in property prediction, transfer learning has recently emerged as a promising solution \citep{Weiss2016, Dai2007, Neyshabur2020}. This technique has so far been utilized by pre-training models on large-scale unlabeled data to learn general molecular representations, which are then fine-tuned on relatively small labeled datasets for specific prediction tasks. In the context of ionic liquids (ILs), one recent study has explored this approach by leveraging extensive unlabeled SMILES data to derive structural embeddings of IL ions, bypassing the need for manually crafted molecular descriptors \citep{Chen2023}. These learned representations were subsequently fine-tuned using limited experimental data to predict various thermophysical properties. While this framework has demonstrated improved generalization and superior performance compared to conventional models, its accuracy notably declines in the presence of extremely sparse experimental data, such as for heat capacity and surface tension. This limitation highlights the need for more robust strategies to ensure reliable predictions across a broader spectrum of IL properties.

In this work, we propose a transfer learning framework that leverages a neural recommender system (NRS) to overcome the limitations posed by sparse experimental data in predicting key physicochemical properties of ILs. The NRS is employed during the pre-training phase to learn property-specific structural embeddings from COSMO-based simulation data for density, viscosity, and heat capacity at fixed temperature and pressure. These embeddings capture essential molecular features without the confounding influence of thermodynamic variables. In the subsequent fine-tuning phase, a simple feedforward neural network is trained for each target property (density, viscosity, surface tension, heat capacity, and melting point) using the learned structural embeddings as inputs and experimental property data for supervision.

This study is organized as follows: In \autoref{sec:methodology}, we detail the methodology, including the neural recommender system and transfer learning framework, the pre-training phase using COSMO-RS simulated data, and the fine-tuning phase with experimental data. This section also presents the datasets, covering the generation of COSMO-based molecular structures, data sampling, and the collection of experimental data from literature. \hyperref[sec:results]{Section~3} discusses the results, evaluating model performance and the impact of the size of the pre-training dataset on prediction accuracy. Finally, \autoref{sec:conclusions} summarizes the conclusions and potential for future application of the proposed methodology.

\section{Methodology}
\label{sec:methodology}

\subsection{Neural recommender system and transfer learning framework}
In this work, we implement a two-step transfer learning framework. An overview of the complete workflow is illustrated in \autoref{fig:workflow}. Separate neural recommender system models are initially pre-trained for each of the three properties: density, viscosity, and heat capacity. This pre-training is conducted using COSMO-generated simulated data at fixed temperature and pressure (\SI{298}{\kelvin} and \SI{1}{\bar}). The goal is to learn property-specific, meaningful structural embeddings for each cation and anion by capturing their interactions and extracting latent features that represent a reduced molecular structural space conditioned on the respective property. These learned embeddings are subsequently fixed and reused, while a feed-forward neural network composed exclusively of dense layers is fine-tuned using sparse experimental data. This setup focuses on modeling the influence of the structural representations on each property under varying temperature and pressure. By sequentially learning structural embeddings and the effects of temperature and pressure, this approach can effectively overcome major challenges in IL property prediction, particularly addressing the limited and imbalanced experimental dataset.

\subsubsection{Pre-training phase}
In the pre-training phase, a neural recommender system is developed to learn property-specific structural embeddings for ILs, as shown in \autoref{fig:nrs_architecture}. Cations and anions are provided as index-based inputs, which are mapped to the target property at fixed temperature and pressure (\SI{298}{\kelvin} and \SI{1}{\bar}) to learn latent vectors of dimension 100, representing structural embeddings for each ion. The embedding vectors for cation and anion are independently processed through a dense neural block resembling a residual network (ResNet) architecture \citep{targ2016resnet}. Each block consists of a series of dense layers with varying number of neurons selected from [100, 200, 300], followed by dropout (rate = 0.05) and element-wise addition layers. The outputs of the individual cation and anion pathways are then concatenated to form a combined latent representation. The concatenated vector is subsequently passed through two additional fully connected layers, as one layer is found to be insufficient, the first with a  number of neurons selected from [50, 100, 200, 400], and the second with half the number of neurons of the preceding layer. Rectified Linear Unit (ReLU) activation functions are applied to all dense layers, while the output layer remains activation-free to prevent range constraints on predicted values. The model is trained using the Adam optimizer \citep{kingma2014adam} with a learning rate of 0.001.

Pre-training is carried out separately for three target properties - density, viscosity, and heat capacity - using identical model architectures and optimization strategies for each property. Hyperparameters related to the number of neurons are tuned individually per property. Pre-training is not performed for surface tension as no corresponding COSMO-RS model is available. Melting point data is also excluded from the pre-training phase since numerous values are missing in the COSMO-RS-based simulation dataset. These gaps primarily arise from convergence issues during the quantum chemical optimization of certain cation–anion structures, which are required for generating reliable COSMO descriptors. To avoid data leakage, it is ensured that ILs included in any of the experimental datasets for the five properties are excluded from all pre-training datasets. Furthermore, rigorous ten-fold cross-validation is conducted during pre-training to select optimal hyperparameters. In each fold, train-test splits are performed rigorously based on ILs rather than randomly on individual data points, ensuring that no ionic liquid appears in both training and test sets.

\subsubsection{Fine-tuning phase}
In the fine-tuning phase, the pre-trained components of the neural recommender system corresponding to cation and anion representations, up to and including the concatenation layer, are reused with fixed weights and treated as a non-trainable forward pass, as shown in \autoref{fig:nrs_architecture}. These embeddings remain constant during fine-tuning. Additionally, temperature and/or pressure inputs are concatenated with the fixed ion embeddings to form the final input representation. This combined vector is then passed through two fully trainable dense layers. The first layer uses a number of neurons selected from [50, 100, 200], while the second layer contains half the number of neurons as the preceding layer. A dropout rate of 0.05 is applied after each dense layer to mitigate overfitting. The Adam optimizer is employed for training with a learning rate of 0.01.

By leveraging the structural knowledge transferred from the pre-trained embeddings, which are learned using well-sampled and unbiased COSMO-generated data, the fine-tuning phase focuses exclusively on modeling the influence of temperature and pressure using sparse and biased experimental datasets. This sequential learning strategy enables effective generalization despite the limited and uneven distribution of experimental data for each property. 10-fold cross-validation is also performed during fine-tuning. Again, to avoid data leakage train-test splits in each fold are performed rigorously based on ionic liquids rather than randomly on individual data points, ensuring that no ionic liquid appears in both training and test sets.

The structural embeddings learned from pre-trained models for the three properties, namely density, viscosity, and heat capacity, are each used as a basis for fine-tuning of all five target properties, including surface tension and melting point. This approach enables the evaluation of both within-property and cross-property transfer learning. The performance outcomes across different source–target property combinations are systematically compared.

As illustrated in Figure 3, the paucity of experimental data highlights the challenges in property prediction and emphasizes the necessity of the transfer learning approach. This methodology enables extrapolation across the entire chemical space of 2,268 cations and 311 anions included in the study. By learning structural embeddings from unbiased simulated data during pre-training, the fine-tuned model not only extrapolates to ILs unseen during training but also predicts properties for combinations where both the cation and the anion are not contained in from the experimental data.

\subsection{Datasets}
\label{sec:datasets}
\subsubsection{Generation of molecular structures via DFT calculations}
First, all unique cations and anions present in the experimental datasets for the five target properties (as listed in \autoref{tab:performance_comparison_1}) are identified. Canonical SMILES strings for each ion are generated using RDKit \citep{landrum2013rdkit}. These SMILES strings are then used to create optimized molecular geometries and COSMO files via a series of quantum chemical calculations. All Density Functional Theory (DFT) calculations are carried out using TURBOMOLE, version 7.7.1 \citep{Furche2014}. Chemical structures are optimized with the BP86 GGA (generalized gradient approximation) functional \citep{Becke1988, Perdew1986}, including DFT-D3 \citep{Grimme2010}, dispersion corrections with Becke-Johnson (BJ) damping \citep{Grimme2011}, and using Ahlrichs’ triple-$\zeta$ valence polarization basis set (def2-TZVP) \citep{Schafer1994} for all atoms.

The input files for COSMO-RS (conductor like screening model for real solvents) \citep{Klamt1995} are generated via single point calculations at BP86-D3(BJ)/def2-TZVPD level of theory, setting an infinite permittivity and using the refined COSMO cavity construction algorithm. The resulting files form the structural input required for property prediction via COSMO-RS.

\subsubsection{Data sampling and COSMO simulations}
After generating structural data for 2,268 cations and 311 anions, a stratified random sampling approach is applied. 50 anions are randomly selected for each cation, leading to a total of approximately 111{,}900 ionic liquids—about 15\% of all 700{,}000+ possible cation-anion combinations considered in this study. We then run simulations to generate pre-training data for these ILs. COSMOtherm 2020 with TZVP parametrization is used to predict three properties: density \citep{Preiss2009}, viscosity \citep{Eiden2011}, and heat capacity \citep{Preiss2009} at fixed temperature (\SI{298}{\kelvin}) and/or pressure (\SI{1}{\bar}). Each property is computed using predefined models available in COSMO-RS, based on the parameter sets reported in the respective references.

\subsubsection{Experimental datasets}
For the fine-tuning step, experimental data for five IL properties are collected from published sources \citep{Paduszynski2019a, Paduszynski2019b, Venkatraman2019, Low2020}, concerning density (2,261 ILs), viscosity (1,995 ILs), surface tension (330 ILs), heat capacity (236 ILs), and melting point (2,178 ILs). After compiling all unique cations and anions across the datasets, data sparsity is evaluated. \autoref{fig:data_sparsity} presents the total number of unique cations (2,268) and anions (311) included in the study, alongside the number of ionic liquids for which experimental data are available for each property. It is evident from the figure that the experimental datasets are highly sparse in comparison to the large number of possible cation-anion combinations. The total number of unique ILs across all properties is influenced by the overlap of cations and anions, with the highest coverage seen in density and melting point data, while surface tension and heat capacity exhibit significantly lower representation due to the limited number of unique ILs (330 and 236, respectively) compared to the total pool of 2,268 cations and 311 anions.

\section{Results and discussion}
\label{sec:results}
\subsection{Model performance and transfer learning outcomes}
The performance of the neural recommender system (NRS) and transfer learning framework developed in this study is summarized in \autoref{tab:performance_comparison_1}. This includes the pre-trained models and the corresponding fine-tuned models for each property. Model effectiveness is evaluated using three standard metrics: coefficient of determination ($R^2$), mean absolute error (MAE), and mean absolute percentage error (MAPE). All scores are calculated using unscaled predicted and target values to maintain consistency and enable comparison with other existing best models in literature.

A ten-fold cross-validation strategy is applied during both the pre-training and fine-tuning phases for hyperparameter optimization. Each fold involves a rigorous split, ensuring that ionic liquids used in the training set are not present in the validation set. Furthermore, all ionic liquids from the five experimental datasets used for fine-tuning are entirely excluded from the pre-training data to avoid any possibility of data leakage.

The fine-tuned models are compared with the best-performing models previously reported in the literature. As shown in \autoref{tab:performance_comparison_1}, the proposed framework achieves significant improvements for properties influenced by both molecular structure as well as temperature and pressure. For example, MAE improvements of 38\% for density, 57\% for viscosity, 75\% for heat capacity, and 43\% for surface tension are observed. These results confirm the benefit of the sequential learning approach proposed in this work. On the other hand, a decrease of 20\% in model performance is observed for the melting point, which depends only on molecular structure. This indicates that transfer learning may not offer as many benefits for structure-only properties. It is worth noting that the experimental dataset for viscosity, expressed as the natural logarithm of viscosity in mPa·s, ranges from 0 to 15. In contrast, the COSMO-RS simulated dataset at a fixed temperature of 298 K spans a wider range of 0 to 0.02 Pa·s, with some ionic liquids showing unrealistically high ln(viscosity) values of up to 0.08 Pa·s. These outliers, likely due to limitations in structure optimization or predictive models, indicate potential noise in the data, as they do not align well with other data during pre-training. Importantly, these noisy data points were intentionally retained to test our strategy’s ability to handle such noise, and it was found that they did not significantly affect the performance of the final fine-tuned model.

To explore the ability to transfer knowledge across different properties, the pre-trained models for density, viscosity, and heat capacity are each used to fine-tune models for all five target properties. These models are trained on a dataset of 113,400 ILs, representing approximately 15\% of all 700{,}000+ possible combinations, all simulated at \SI{298}{\kelvin} using COSMO calculations. As illustrated in \autoref{fig:exp_data}, the density model is used as a base to fine-tune all five property models. The performance metrics for each case, including $R^2$, MAE, and MAPE, are presented alongside the results. The results indicate that the structural representations learned by the density model enhance the prediction accuracy for four out of the five properties compared to existing methods.

\autoref{fig:density_transfer} further demonstrates the effectiveness of the viscosity and heat capacity models as sources of transferable knowledge. Small performance improvements are observed when the same property is used for both pre-training and fine-tuning. For instance, the viscosity model achieves an MAE of 0.0729 when also pre-trained on viscosity data, compared to 0.0759 when a model pre-trained on heat capacity is applied. Similarly, the heat capacity model achieves an MAE of 0.1515 in the within-property transfer case, slightly outperforming the MAE of 0.1599 when transferring knowledge from viscosity data. To illustrate the model’s predictive capabilities, \autoref{fig:visc_hc_transfer} shows the predicted versus experimental values of each property as a function of temperature and/or pressure for a representative ionic liquid using the best fine-tuned models. This further validates the model’s ability to capture complex relationships with high fidelity.

The results demonstrate that the proposed transfer learning framework can successfully capture and transfer structural knowledge. This approach is particularly useful for predicting properties that depend on molecular structure as well as temperature and pressure. However, for properties dependent only on structure such as melting point, transferring such embeddings results in a decline in predictive performance compared to current benchmark. This may reflect the fact that the benchmark \citep{Chen2023} utilizes more structural information compared to the proposed approach. The transfer learning framework effectively mitigates biases in experimental data, particularly for properties like surface tension and heat capacity, where experimental datasets include only a small subset of ionic liquids, ranging from 64 (surface tension) to 2,261 (density) ILs depending on the property. This approach enables property prediction across the entire combinatorial space of 705,348 ionic liquids formed from 2,268 cations and 311 anions, showcasing its ability to extrapolate within a wide chemical space with limited experimental data. However, deviations in predictive accuracy are observed at the extreme ends of the heat capacity value range, occurring in both within-property and cross-property knowledge transfer during fine-tuning, as shown in \autoref{fig:density_transfer}. This indicates that inter-property knowledge transfer is not the cause of the deviation, reinforcing the impact of sparse experimental data. Although the framework surpasses existing benchmarks for these properties, expanding experimental datasets for surface tension and heat capacity with additional ionic liquids is recommended to further enhance model performance and improve generalization across the chemical space.

The well-established interdependence of surface tension ($\sigma$), density ($\rho$), and dynamic viscosity ($\mu$) is supported by prior studies, which link these properties through molecular interactions and fluid characteristics \citep{Sugden1924, Singh2011, Oroian2013}. \cite{Sugden1924} quantified the relationship between surface tension and density via the parachor $P$, a molecular descriptor reflecting chemical composition (see \hyperref[eq:parachor]{Eq.~1}).  \cite{Singh2011} extended these correlations to ionic liquids, showing that denser fluids with stronger intermolecular forces exhibit higher viscosity and surface tension, often following empirical relations such as \hyperref[eq:visc-density]{Eq.~2}. \cite{Oroian2013} further validated these interconnections in complex substances like honey, where temperature-dependent correlations indicate that viscosity and surface tension scale with density, as represented in \hyperref[eq:combined]{Eq.~3}. Our framework of transferring knowledge across properties essentially utilizes these correlations.

\begin{align}
\sigma^{1/4} &= k_1 \rho (P), \label{eq:parachor} \\
\mu &= k_2 \rho^a, \label{eq:visc-density} \\
\sigma &= k_3 \mu^b \rho^c, \label{eq:combined}
\end{align}

\subsection{Impact of the size of the pre-training dataset on model performance performance}
To understand how the amount of pre-training data influences model performance, additional calculations are performed. From the full set of 2,268 cations and 311 anions, resulting in 705,348 possible ionic liquids, subsets are created using varying sampling levels. Specifically, one, five, ten, twenty, thirty-five, and fifty anions are sampled per cation, corresponding to approximately 0.3\%, 1.6\%, 3.2\%, 6.4\%, 11.2\%, and 15\% of the total cation-anion combinations. Pre-training is performed on these subsets, and both pre-training and fine-tuning performances are recorded in \autoref{tab:performance_comparison_2}.

The pre-training MAE of the density model decreases significantly from 54.49 to 5.12 as the size of the pre-training dataset increases from 0.3\% (each cation sampled with one anion) to 3.2\%. Further increasing the pre-training data to 15\% reduces the pre-training MAE to 2.83. Fine-tuning performance across all five properties remains similar regardless of the size of the pre-training data used, with approximately 10\% variation in accuracy observed for all properties. This analysis is presented for each property in \autoref{tab:performance_comparison_2} and visualized in \cref{fig:supp_fig1,fig:supp_fig2,fig:supp_fig3,fig:supp_fig4,fig:supp_fig5}
in the Supplementary Information. Notably, increasing COSMO data from 0.3\% ($R^2 = 0.83$) to 1.6\% (each cation sampled with five anions) significantly improves pre-training performance to $R^2 = 0.993$, with further increases yielding only marginal gains. Remarkably, fine-tuning performance for all properties remains comparable even when using a pre-trained density model with only 0.3\% COSMO data and $R^2 = 0.83$, which is modest compared to models with larger COSMO datasets. To validate the effectiveness of the transferred structural knowledge, each fine-tuned model is compared to a random baseline model with fixed weights and an identical feed-forward network architecture, as shown in \cref{fig:supp_fig1,fig:supp_fig2,fig:supp_fig3,fig:supp_fig4,fig:supp_fig5}. The random baseline models exhibit significantly poorer fine-tuning performance, highlighting the meaningful structural knowledge transferred by the proposed strategy. Even the model pre-trained on only 0.3\% of ILs achieves fine-tuning accuracy comparable to models with larger datasets, surpassing all benchmarks. This underscores the power of the proposed transfer learning framework, demonstrating its ability to leverage minimal pre-training data for robust property prediction. The models pre-trained on density are further fine-tuned on all properties. A similar impact of the amount of COSMO data is observed in the pre-training performance of viscosity and heat capacity, (\cref{fig:supp_fig6}). Although not shown, fine-tuning performance follows the same trend for all properties when using pre-trained viscosity or heat capacity models.

\section{Conclusions}
\label{sec:conclusions}
In this study, we integrate a neural recommender system (NRS) in a transfer learning framework to predict multiple properties of ionic liquids using limited experimental data. The input features are learned sequentially: For molecular structures, we pre-train the NRS models using simulated COSMO data and then fine-tune them on limited experimental datasets to capture the effects of temperature and pressure. The results demonstrate that structural embeddings learned under fixed thermodynamic conditions can be effectively reused to predict properties influenced by both molecular structure as well as temperature and pressure. By efficiently generating COSMO data at fixed temperature and pressure to cover the entire cation-anion space, we ensure unbiased learning of structural features. These unbiased structural representations enable effective fine-tuning on biased experimental data across all temperatures and pressures.

Our approach achieves significant improvements in prediction accuracy for density, viscosity, heat capacity, and surface tension. These improvements are consistent across both within-property and cross-property transfer learning. For surface tension and melting point, for which no simulated data is available, the framework successfully transfers knowledge from models pre-trained on other properties. However, no improvement is observed in the prediction of melting points, indicating that models with access to more information on molecular structure, such as the one proposed by \cite{Chen2023}, achieve superior performance for properties which do not depend on temperature or pressure.

The proposed strategy enables property prediction across a vast chemical space of over 700{,}000 ILs, despite limited availability of experimental data. This highlights the framework's scalability and strong extrapolation capabilities, offering a powerful tool for screening ILs in process design. We provide a ready-to-use Python function available on GitHub (link to be added) to directly input the cation ID, anion ID, temperature, and/or pressure and predict the five properties included in this study.

Future extensions may include adapting the framework to predict additional thermophysical properties. For instance, interaction-based properties such as activity coefficients could be modeled using sequential knowledge transfer from activity coefficients at infinite dilution at fixed temperature to those at varying temperatures, and eventually to finite dilution data. This progression could enable prediction of complete vapor–liquid and liquid–liquid equilibria for solute–IL systems.

\section*{Conflict of interest}
No conflicts of interest have been identified in relation to this work.

\section*{Acknowledgments}
The authors gratefully acknowledge the support of this research work by the European Regional Development Fund (ERDF), provided by the State of Saxony-Anhalt within the programme "Research and Innovation", via the project “Further Development of the Center for Dynamic Systems (CDS) – Subproject OVGU" (grant number: ZS/2023/12/182075). This work is also a part of the Research Cluster 
“SmartProSys: Intelligent Process Systems for the Sustainable Production of Chemicals” funded by the Ministry for Science, Energy, Climate Protection and the Environment of the State of Saxony-Anhalt in Germany. Sahil Sethi is also affiliated with the International Max Planck Research School for Systems and Process Engineering for a Sustainable Chemical Production (IMPRS SysProSus) at Otto von Guericke University Magdeburg. Special recognition is extended to Froze Jameel for his valuable assistance in configuring the COSMO calculations using TURBOMOLE. Additionally, the authors would like to thank their former colleague Edgar Ivan Sanchez Medina for the constructive discussions concerning the methodology.

\section*{Code and data availability}
The source code, training routines, experimental data, trained models, and results from this study will be made publicly available in a designated GitHub repository. The link will be specified upon publication.

\section*{Author contributions}
Sahil Sethi contributed to conceptualization, formal analysis, methodology, software development, and drafting the initial manuscript; Kai Sundmacher was responsible for funding acquisition, supervision, and manuscript review and editing; Caroline Ganzer contributed to conceptualization, methodology, supervision, and manuscript review and editing.

\section*{Use of AI tools}
Generative AI tools were used to assist in refining language, editing codes, and improving clarity in the manuscript. All scientific content, interpretations, and conclusions were developed by the authors.

\section*{Figures}
\ifnum\value{IncludeFigures}=1
\IfFileExists{Figure1.pdf}{%
\begin{figure}[htbp]
\centering
\includegraphics[width=\linewidth]{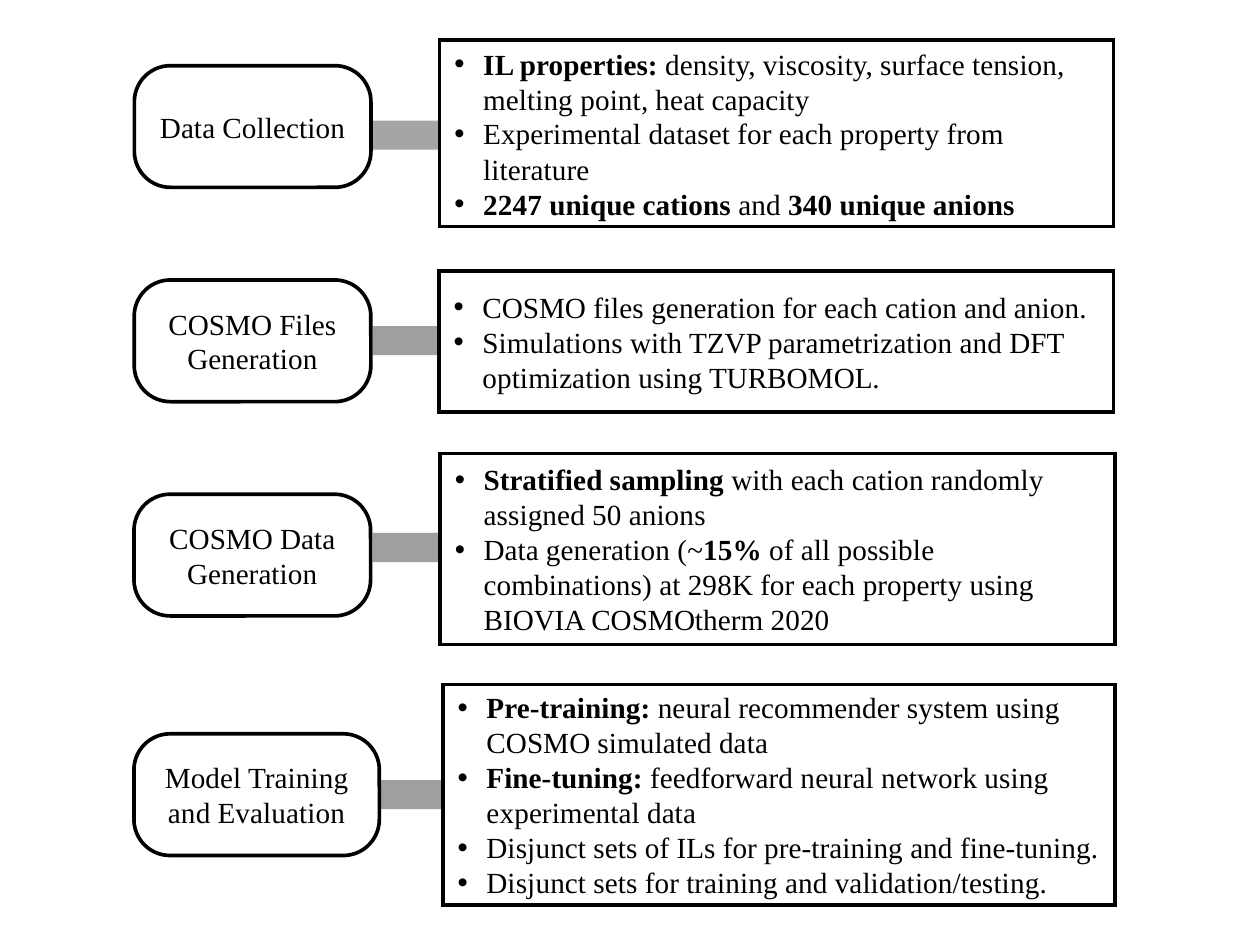}
\caption{Overview of the complete workflow for the proposed transfer learning framework.}
\label{fig:workflow}
\end{figure}
\clearpage
}{}
\IfFileExists{Figure2.pdf}{%
\begin{figure}[htbp]
\centering
\includegraphics[width=\linewidth]{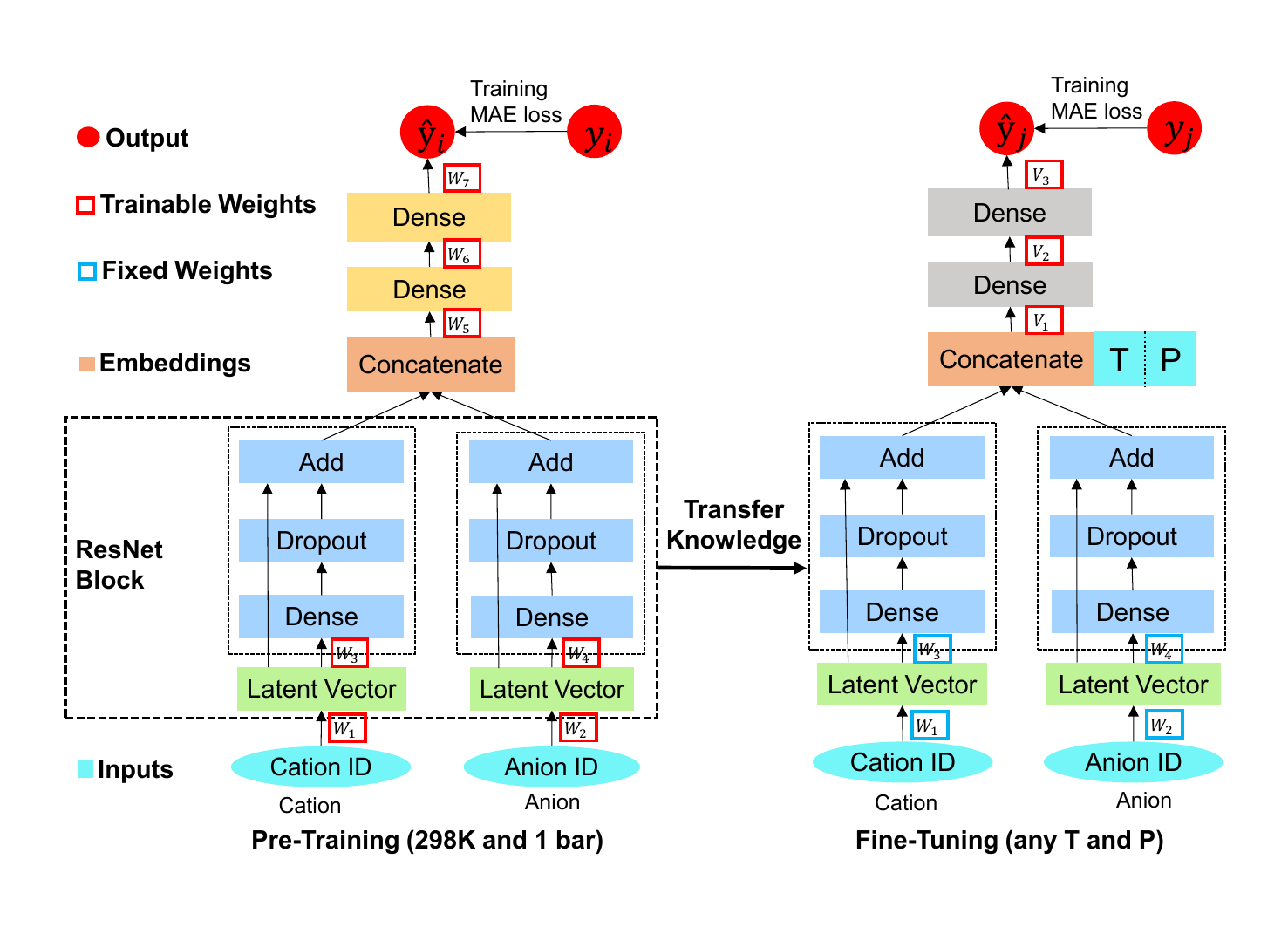}
\caption{Architecture of the proposed framework of transferring knowledge using a neural recommender system for pre-training and a feedforward neural network for fine-tuning.}
\label{fig:nrs_architecture}
\end{figure}
\clearpage
}{}
\IfFileExists{Figure3.pdf}{%
\begin{figure}[htbp]
\centering
\includegraphics[width=\linewidth]{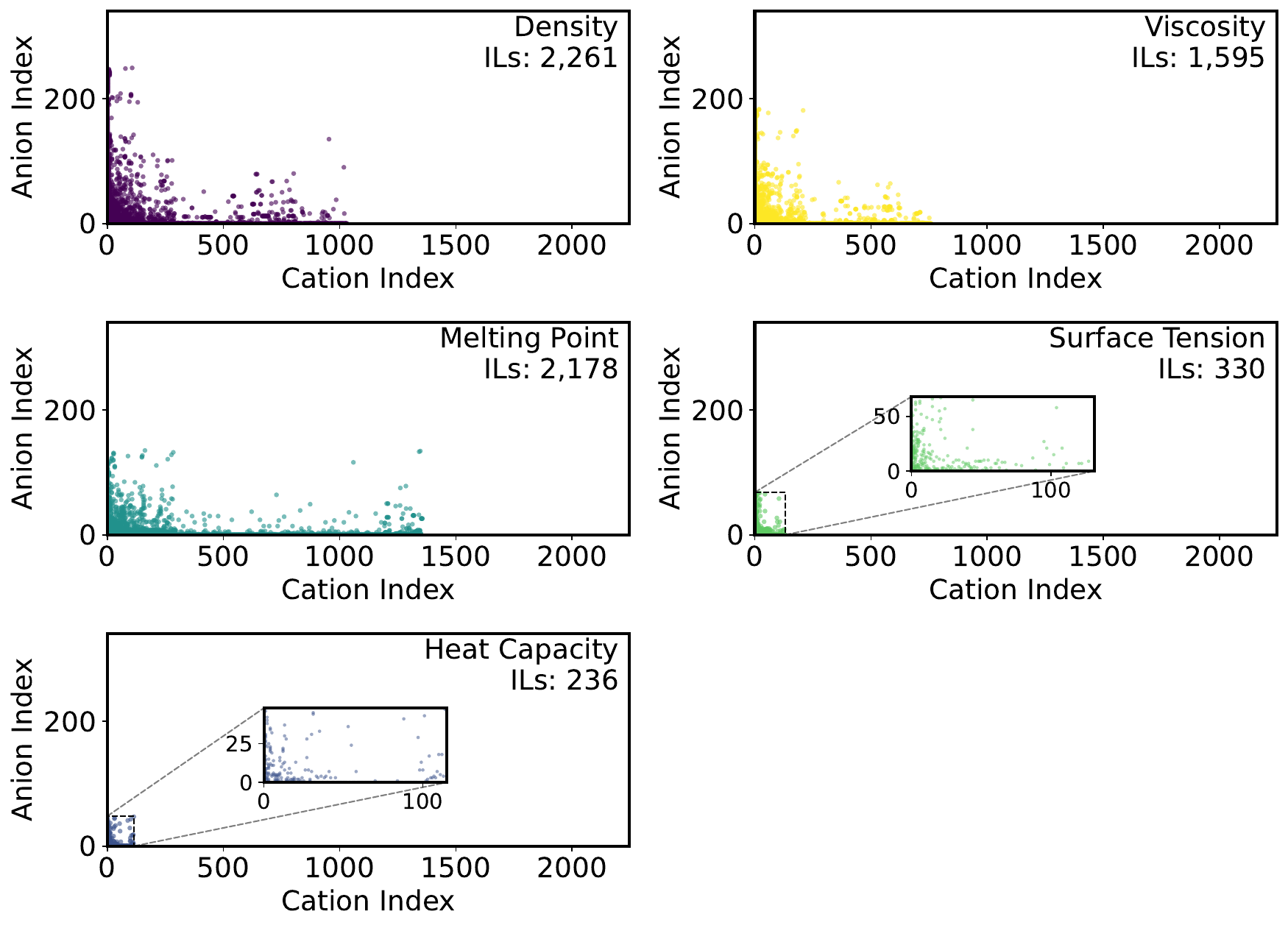}
\caption{Sparsity in the experimental data for the five properties of ionic liquids.}
\label{fig:data_sparsity}
\end{figure}
\clearpage
}{}
\IfFileExists{Figure4.pdf}{%
\begin{figure}[htbp]
\centering
\includegraphics[width=0.8\linewidth]{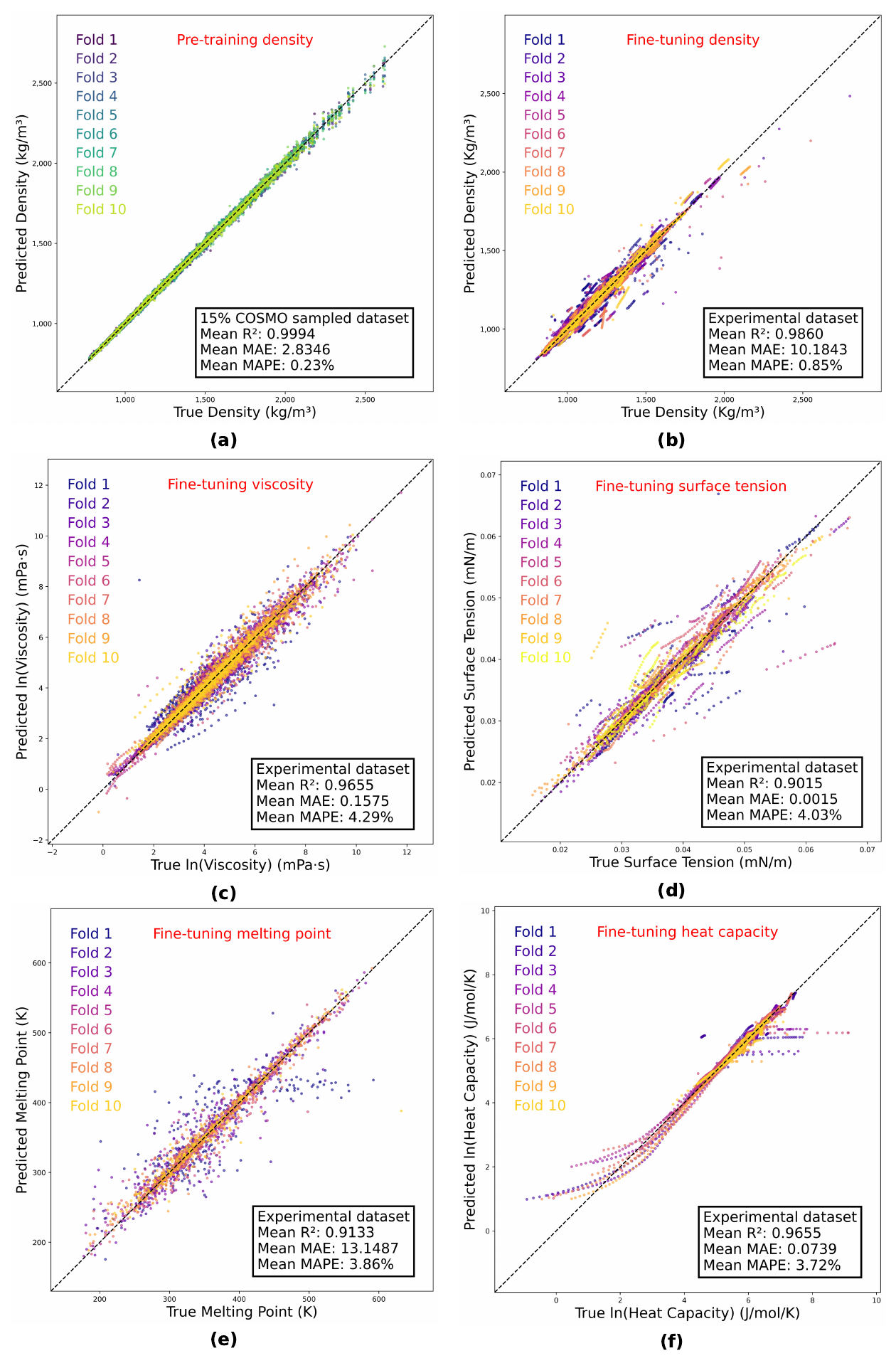}
\caption{Results of the transfer learning framework: performance of the model (a) pre-trained on density, and subsequently fine-tuned on (b) density, (c) viscosity, (d) surface tension, (e) melting point, and (f) heat capacity.}
\label{fig:exp_data}
\end{figure}
\clearpage
}{}
\IfFileExists{Figure5.pdf}{%
\begin{figure}[htbp]
\centering
\includegraphics[width=0.8\linewidth]{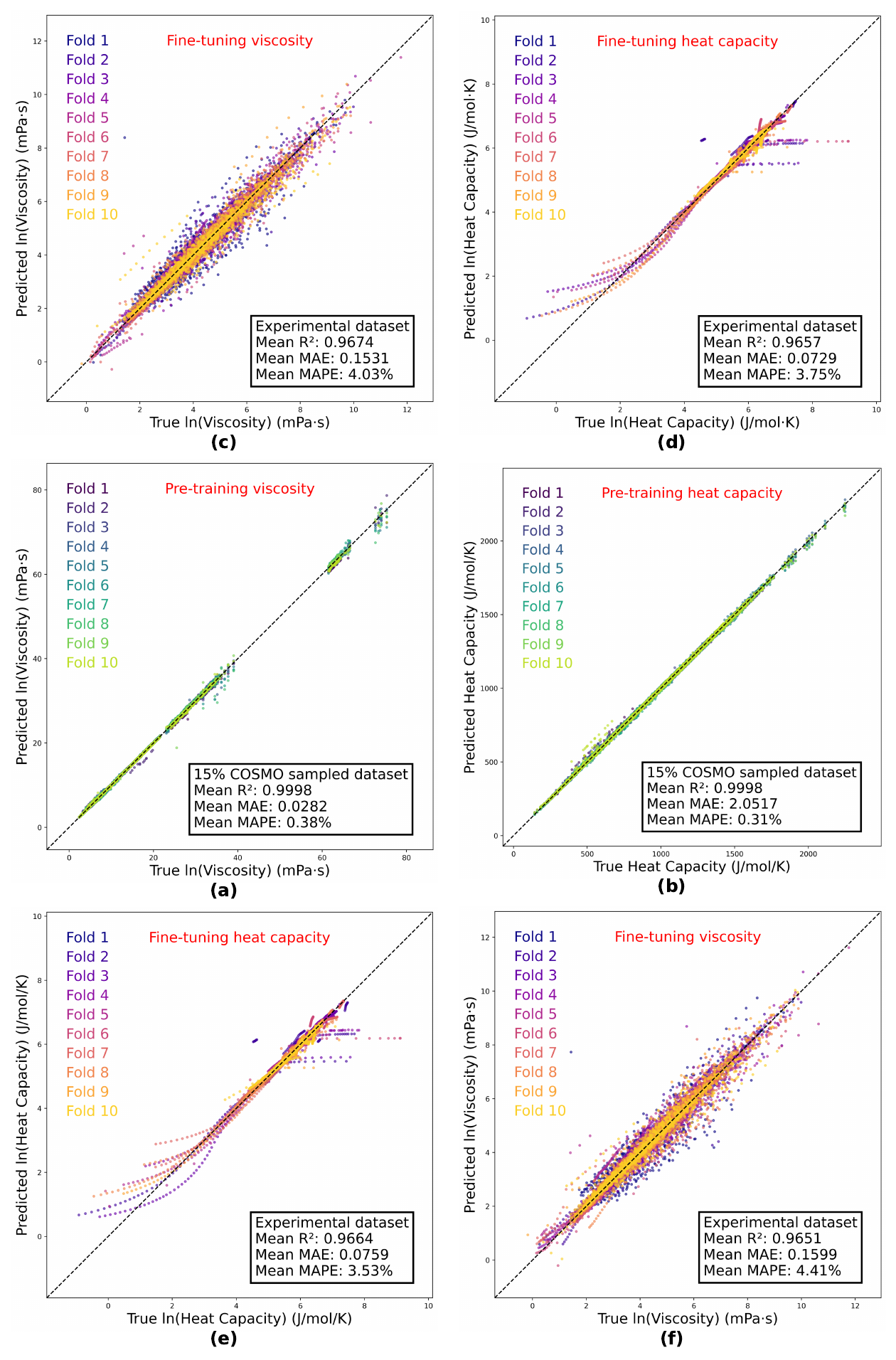}
\caption{Performance analysis of transfer learning: (a) pre-training on viscosity data, then fine-tuning on (c) viscosity and (e) heat capacity; (b) pre-training on heat capacity data, then fine-tuning on (d) heat capacity and (f) viscosity.}
\label{fig:density_transfer}
\end{figure}
\clearpage
}{}
\IfFileExists{Figure6.pdf}{%
\begin{figure}[htbp]
\centering
\includegraphics[width=\linewidth]{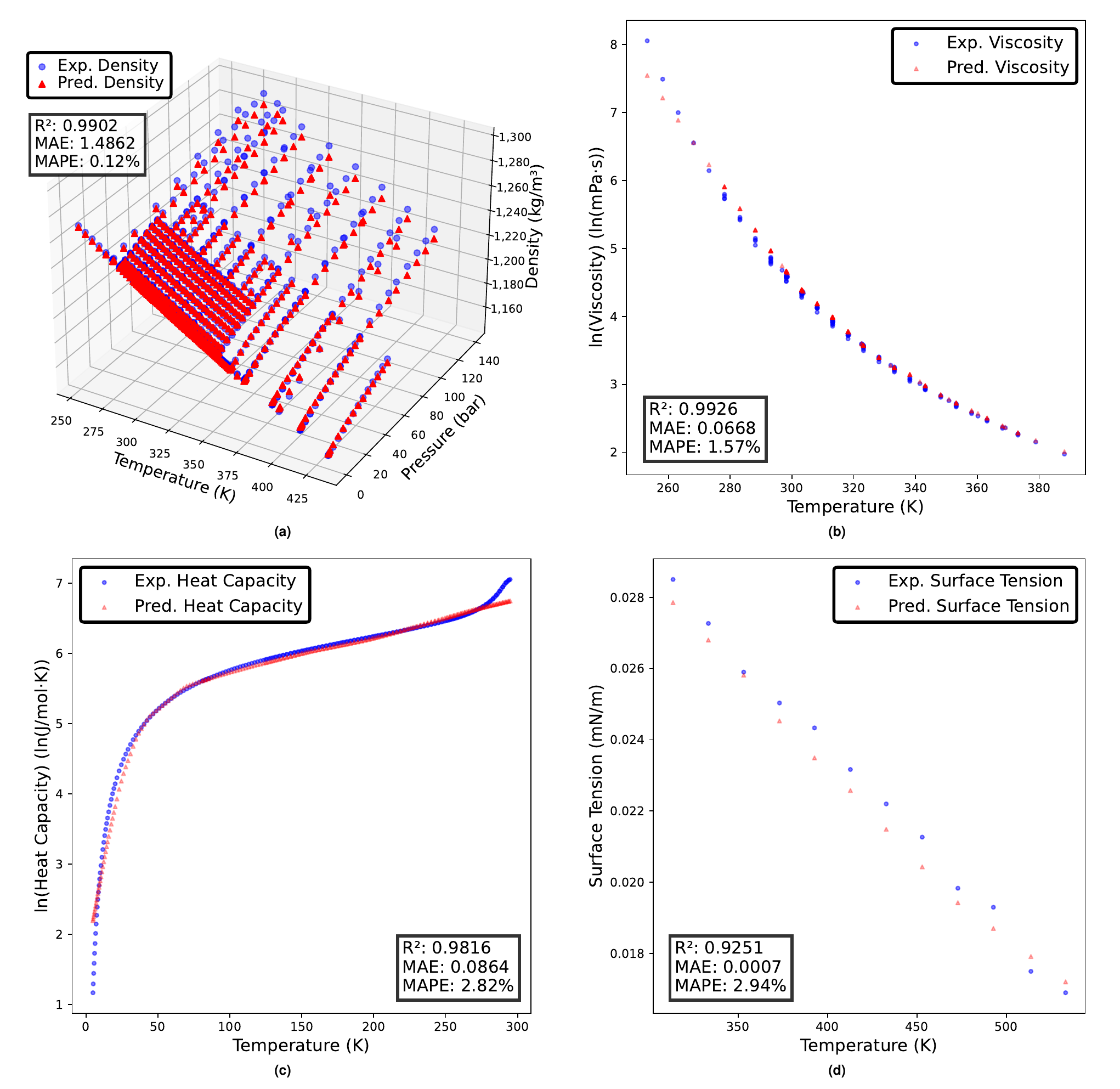}
\caption{Comparison of experimental vs. predicted values for a single ionic liquid using the best-performing model for each property as shown in Figures 4 and 5: (a) density, (b) viscosity, (c) heat
capacity, (d) surface tension.}
\label{fig:visc_hc_transfer}
\end{figure}
\clearpage
}{}
\else
\fi

\section*{Tables}
\ifnum\value{IncludeTables}=1
\IfFileExists{table1.tex}{%
\input{table1.tex}
\clearpage
}{}
\IfFileExists{table2.tex}{%
\input{table2.tex}
\clearpage
}{}
\else
\fi

\section*{Supplementary Figures}
\ifnum\value{IncludeSupplementaryFigures}=1
\setcounter{figure}{0}
\renewcommand{\thefigure}{S\arabic{figure}}
\IfFileExists{Figure_S1.pdf}{%
\begin{figure}[htbp]
\centering
\includegraphics[width=\linewidth]{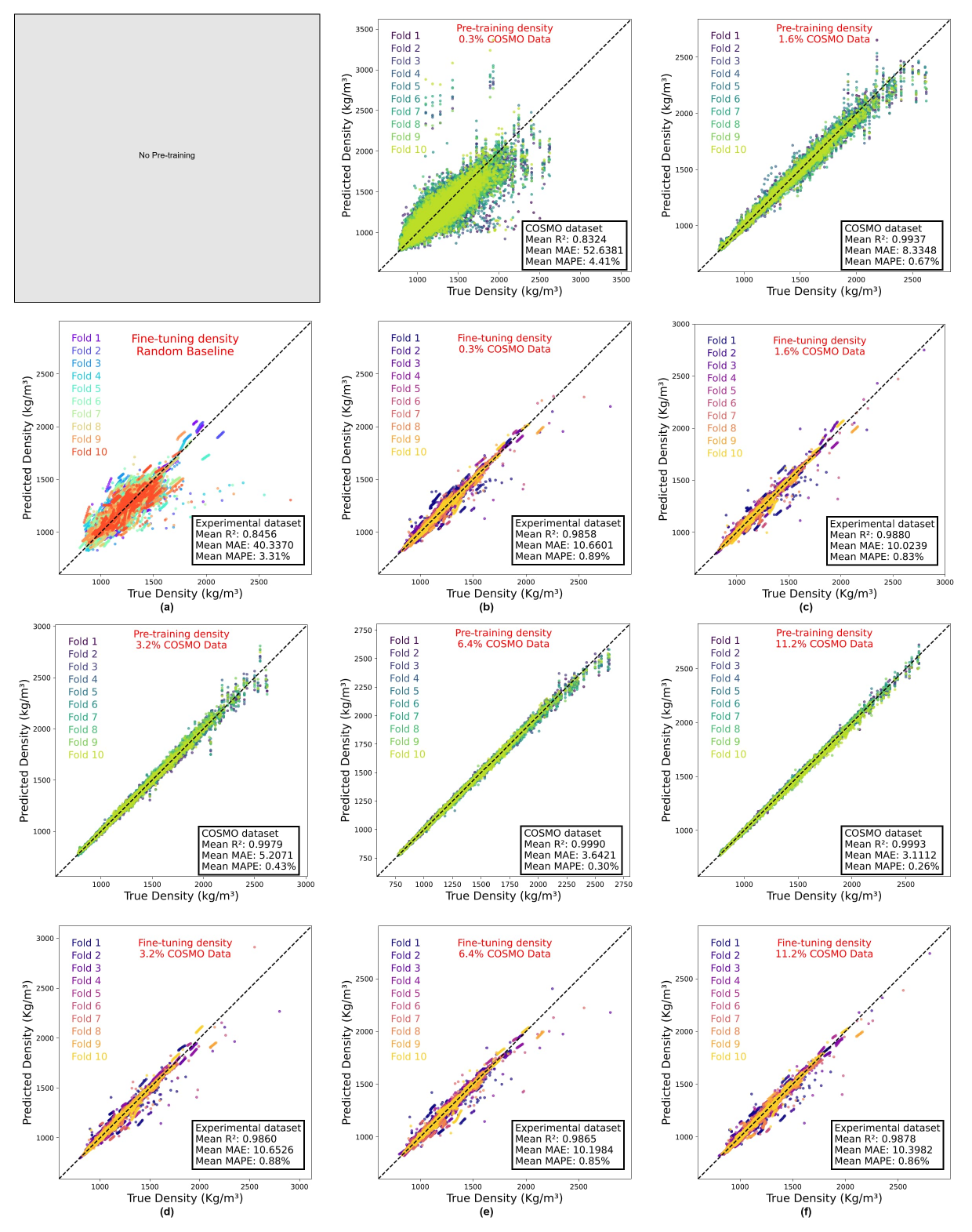}
\caption{Impact of the amount of pre-training data on model performance for pre-training density and fine-tuning density.}
\label{fig:supp_fig1}
\end{figure}
\clearpage
}{}
\IfFileExists{Figure_S2.pdf}{%
\begin{figure}[htbp]
\centering
\includegraphics[width=\linewidth]{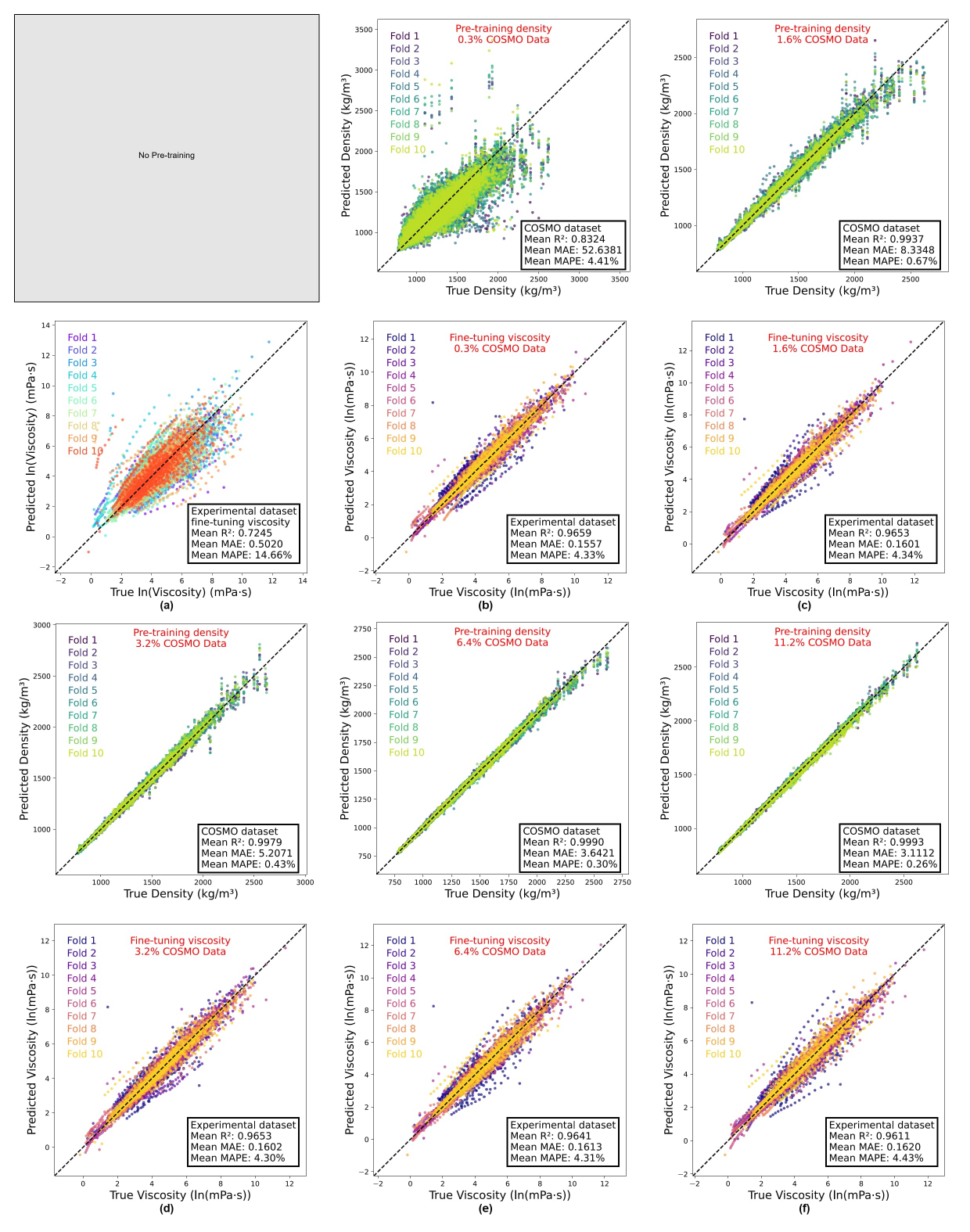}
\caption{Impact of the amount of pre-training data on model performance for pre-training density and fine-tuning viscosity.}
\label{fig:supp_fig2}
\end{figure}
\clearpage
}{}
\IfFileExists{Figure_S3.pdf}{%
\begin{figure}[htbp]
\centering
\includegraphics[width=\linewidth]{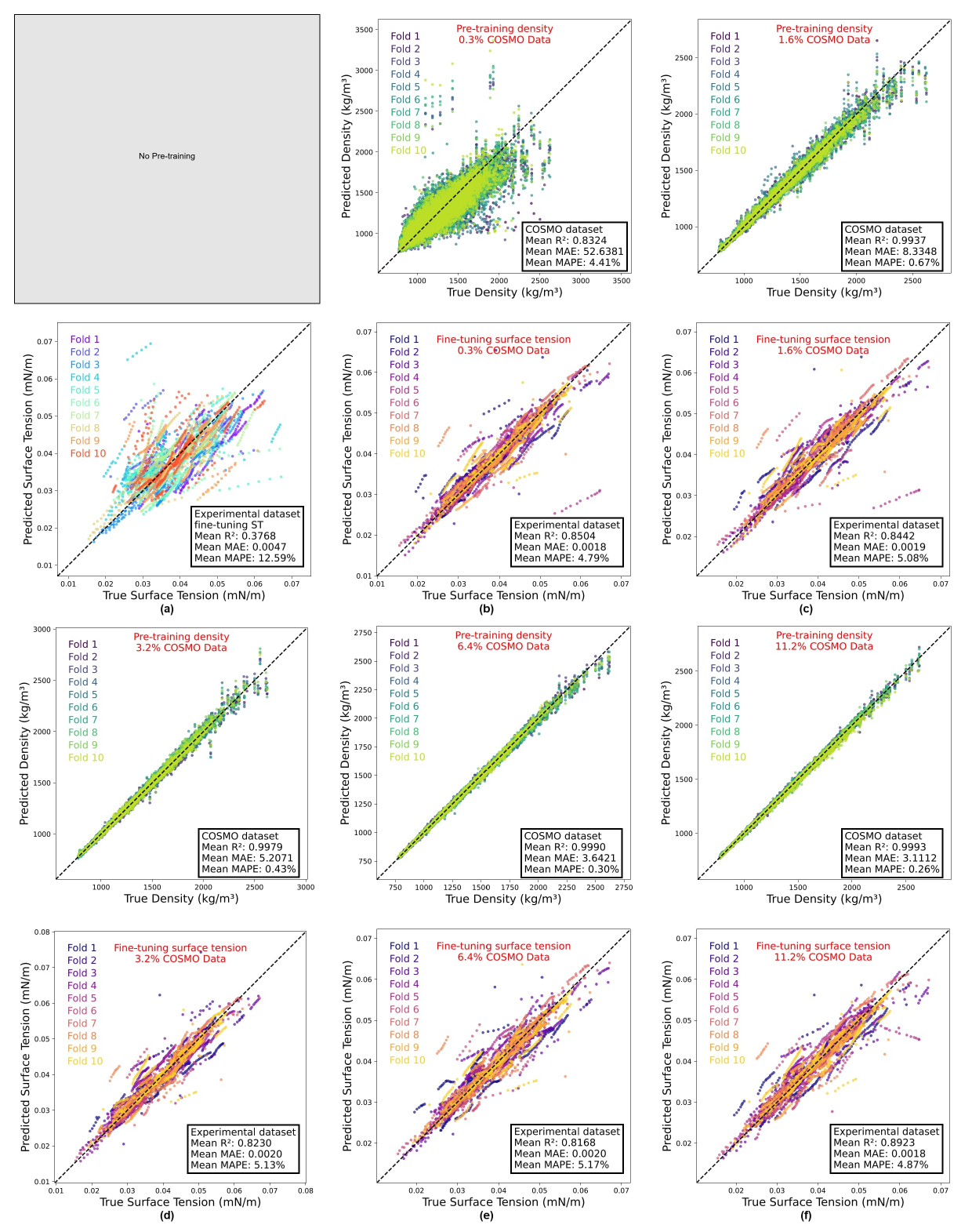}
\caption{Impact of the amount of pre-training data on model performance for pre-training density and fine-tuning surface tension.}
\label{fig:supp_fig3}
\end{figure}
\clearpage
}{}
\IfFileExists{Figure_S4.pdf}{%
\begin{figure}[htbp]
\centering
\includegraphics[width=\linewidth]{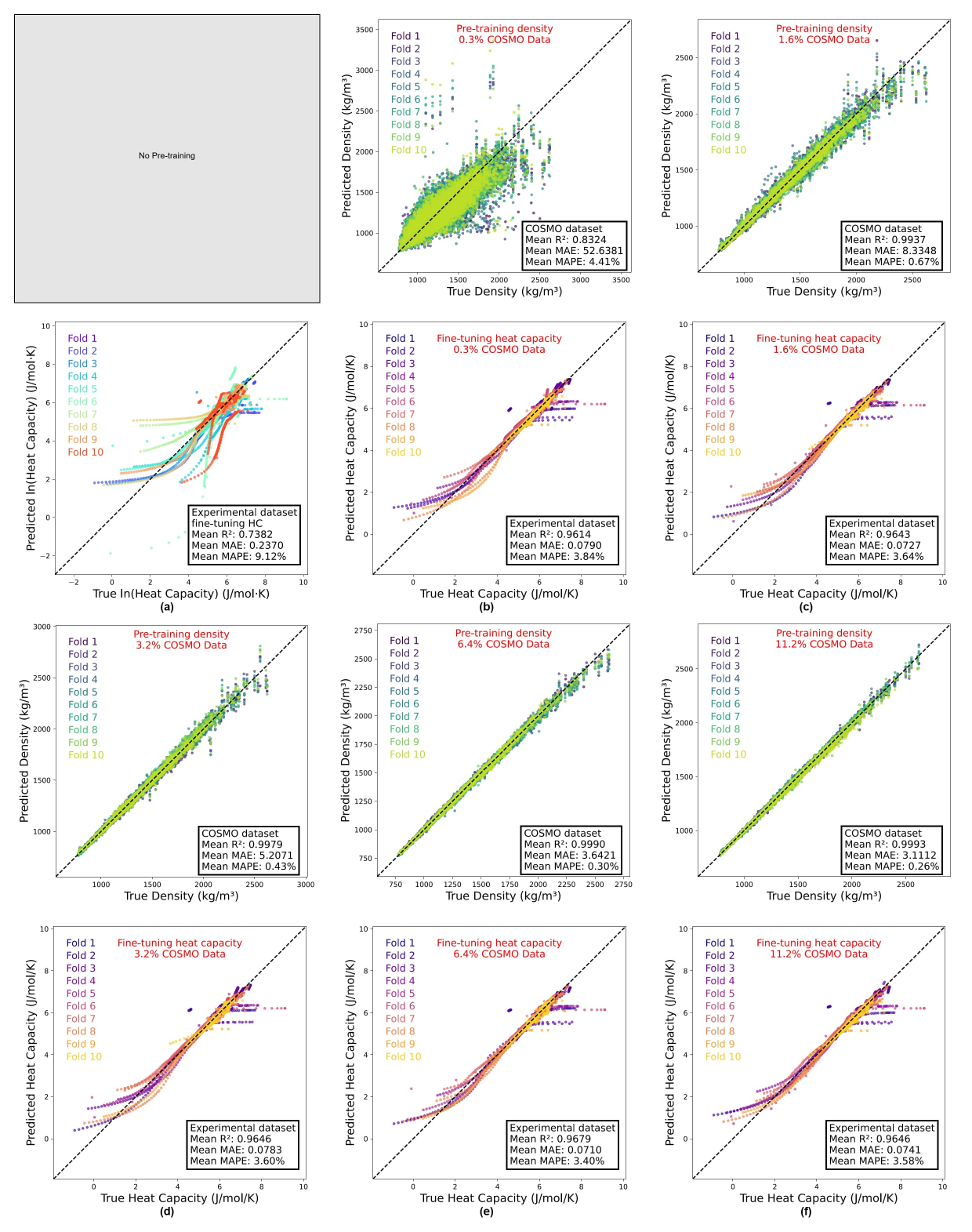}
\caption{Impact of the amount of pre-training data on model performance for pre-training density and fine-tuning heat capacity.}
\label{fig:supp_fig4}
\end{figure}
\clearpage
}{}
\IfFileExists{Figure_S5.pdf}{%
\begin{figure}[htbp]
\centering
\includegraphics[width=\linewidth]{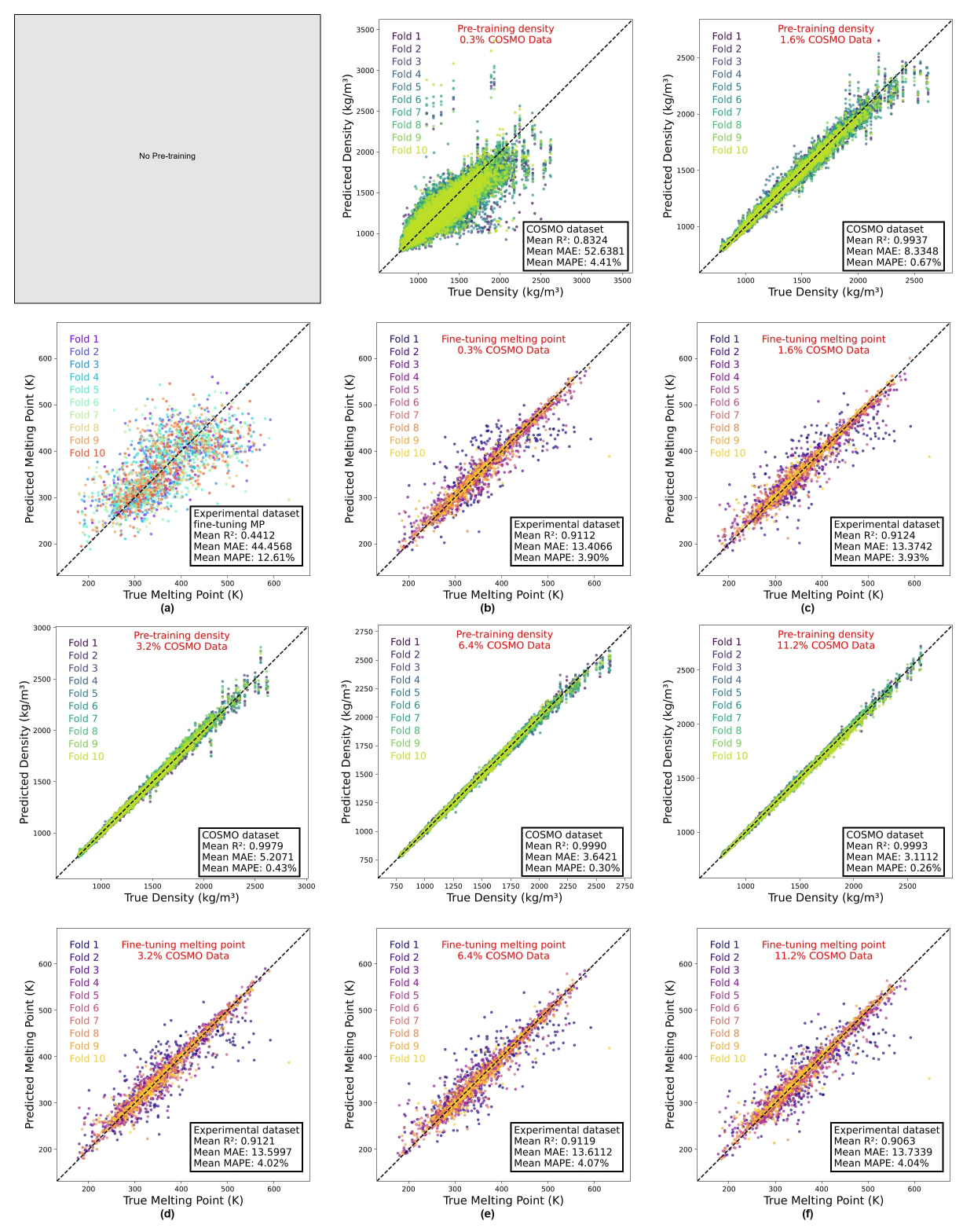}
\caption{Impact of the amount of pre-training data on model performance for pre-training density and fine-tuning melting point.}
\label{fig:supp_fig5}
\end{figure}
\clearpage
}{}
\IfFileExists{Figure_S6.pdf}{%
\begin{figure}[htbp]
\centering
\includegraphics[width=\linewidth]{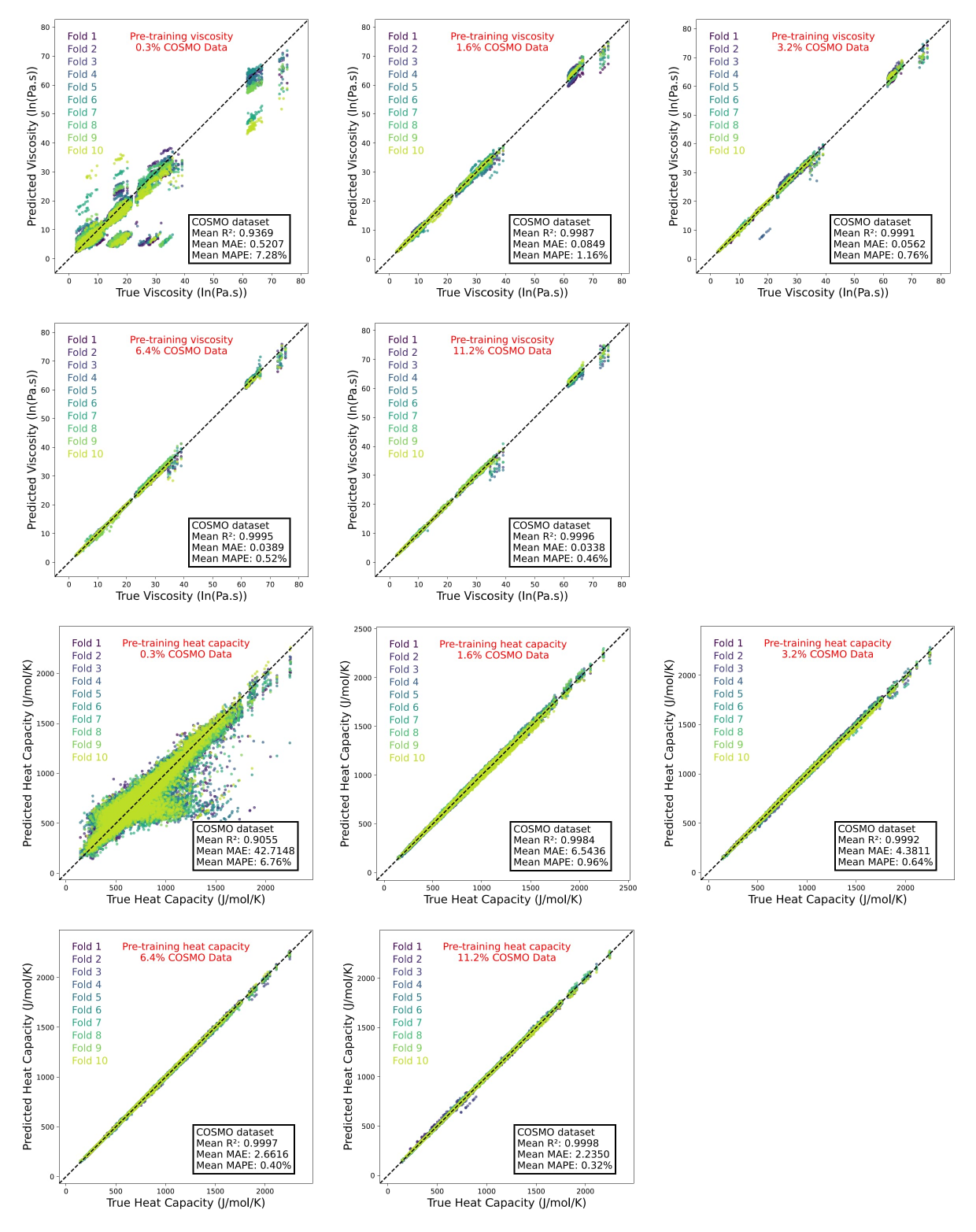}
\caption{Impact of the amount of pre-training data on model performance for pre-training viscosity and heat capacity.}
\label{fig:supp_fig6}
\end{figure}
\clearpage
}{}
\clearpage
\renewcommand{\thefigure}{\arabic{figure}}
\fi

\clearpage

\bibliography{references} 

\end{document}

%% file: table1.tex
\begin{sidewaystable}[h!]
\centering
\caption{Comparative performance of the proposed approach compared to existing benchmarks for the five IL properties examined in this study.}
\label{tab:performance_comparison_1}
\begin{adjustbox}{max width=\textwidth}
\begin{tabular}{lccccccccccc}
\toprule
 & \multicolumn{4}{c}{Experimental dataset} & \multicolumn{4}{c}{COSMO dataset} & Method & Test MAE (split by IL) & Source \\
\cmidrule(lr){2-5} \cmidrule(lr){6-9}
& \# ILs & \# unique cations & \# unique anions & & \# ILs & \# unique cations & \# unique anions & & & & \\
\midrule
Density & 2,261 & 1,032 & 250 & & 108,093 & 2,209 & 311 & & LSSVM & -- & \cite{Paduszynski2019a} \\
& & & & & & & & & ILTrans & 16.46 & \cite{Chen2023} \\
& & & & & & & & & This work & 10.18 & \\
\midrule
Viscosity & 1,595 & 760 & 184 & & 108,093 & 2,209 & 311 & & LSSVM & -- & \cite{Paduszynski2019b} \\
& & & & & & & & & ILTrans & 0.35 & \cite{Chen2023} \\
& & & & & & & & & This work & 0.15 & \\
\midrule
Surface tension & 330 & 131 & 68 & & & & & & GBM & -- & \cite{Venkatraman2019} \\
& & & & & & & & & ILTrans & 0.0030 & \cite{Chen2023} \\
& & & & & & & & & This work & 0.0017 & \\
\midrule
Heat capacity & 236 & 115 & 48 & & 108,093 & 2,209 & 311 & & GBM & -- & \cite{Venkatraman2019} \\
& & & & & & & & & ILTrans & 0.28 & \cite{Chen2023} \\
& & & & & & & & & This work & 0.07 & \\
\midrule
Melting point & 2,178 & 1,357 & 136 & & & & & & KRR & 29.78 & \cite{Low2020} \\
& & & & & & & & & ILTrans & 11.15 & \cite{Chen2023} \\
& & & & & & & & & This work & 13.44 & \\
\bottomrule
\end{tabular}
\end{adjustbox}
\end{sidewaystable}

%% file: table2.tex
\begin{sidewaystable}
\centering
\caption{Performance of the proposed neural recommender system with transfer learning for IL property prediction for varying size of the pre-training data set.}
\label{tab:performance_comparison_2}
\begin{adjustbox}{max width=\textwidth}
\begin{tabular}{cc c c cc cc cc cc}
\toprule
\multicolumn{2}{c}{\textbf{Property}} & \multirow{2}{*}{\textbf{Metric}} 
& \textbf{Random baseline} 
& \multicolumn{2}{c}{\textbf{0.3\% COSMO data}} 
& \multicolumn{2}{c}{\textbf{1.6\% COSMO data}} 
& \multicolumn{2}{c}{\textbf{11.2\% COSMO data}} 
& \multicolumn{2}{c}{\textbf{15\% COSMO data}} \\
\cmidrule(r){1-2} \cmidrule(lr){4-4} \cmidrule(lr){5-6} \cmidrule(lr){7-8} \cmidrule(lr){9-10} \cmidrule(l){11-12}
\textbf{Pre-train} & \textbf{Fine-tune} & & \textbf{Fine-tune} 
& \textbf{Pre-train} & \textbf{Fine-tune} 
& \textbf{Pre-train} & \textbf{Fine-tune} 
& \textbf{Pre-train} & \textbf{Fine-tune} 
& \textbf{Pre-train} & \textbf{Fine-tune} \\
\midrule
\multirow{2}{*}{Density} & \multirow{2}{*}{Density} 
& MAE/MAPE & 40.33 / 3.31\% & 52.63 / 4.41\% & 10.66 / 0.89\% & 8.33 / 0.67\% & 10.02 / 0.83\% & 3.11 / 0.26\% & 10.40 / 0.86\% & 2.83 / 0.23\% & 10.18 / 0.85\% \\
& & $R^2$     & 0.845 & 0.832          & 0.986          & 0.994         & 0.988          & 0.999          & 0.988           & 0.999 & 0.986 \\
\multirow{2}{*}{Density} & \multirow{2}{*}{Viscosity} 
& MAE/MAPE & 0.502 / 14.66\%  & 52.63 / 4.41\% & 0.156 / 4.33\% & 8.33 / 0.67\% & 0.160/ 4.34\% & 3.11 / 0.26\% & 0.162/ 4.43\% & 2.83 / 0.23\% & 0.158/ 4.29\% \\
& & $R^2$     & 0.845 & 0.832          & 0.986          & 0.994         & 0.988          & 0.999          & 0.988           & 0.999 & 0.986 \\

\multirow{2}{*}{Density} & \multirow{2}{*}{Surface Tension} 
& MAE/MAPE & 0.005 / 12.59\% & 52.63 / 4.41\% & 0.002 / 4.79\% & 8.33 / 0.67\% & 0.002 / 5.08\% & 3.11 / 0.26\% & 0.002 / 4.87\% & 2.83 / 0.23\% & 0.002 / 4.49\% \\
& & $R^2$     & 0.377 & 0.832          & 0.850          & 0.994         & 0.844         & 0.999          & 0.892           & 0.999 & 0.869 \\

\multirow{2}{*}{Density} & \multirow{2}{*}{Heat Capacity} 
& MAE/MAPE & 0.237 / 9.12\% & 52.63 / 4.41\% & 0.079 / 3.84\% & 8.33 / 0.67\% & 0.073 / 3.64\% & 3.11 / 0.26\% & 0.074 / 3.58\% & 2.83 / 0.23\% & 0.074 / 3.72\% \\
& & $R^2$     & 0.738 & 0.832          & 0.964          & 0.994         & 0.964          & 0.999          & 0.965           & 0.999 & 0.966 \\

\multirow{2}{*}{Density} & \multirow{2}{*}{Melting Point} 
& MAE/MAPE & 44.46 / 12.61\% & 52.63 / 4.41\% & 13.41 / 3.90\% & 8.33 / 0.67\% & 13.37 / 3.93\% & 3.11 / 0.26\% & 13.73 / 4.04\% & 2.83 / 0.23\% & 13.15 / 3.86\% \\
& & $R^2$     & 0.441 & 0.832          & 0.911         & 0.994         & 0.912          & 0.999          & 0.906           & 0.999 & 0.966 \\
\bottomrule
\end{tabular}
\end{adjustbox}
\end{sidewaystable}